# Genetic Algorithm for a class of Knapsack Problems

Shalin Shah
shalin@alumni.usc.edu

**Abstract**

The 0/1 knapsack problem is weakly NP-hard in that there exist pseudo-polynomial time algorithms based on dynamic programming that can solve it exactly. There are also the core branch and bound algorithms that can solve large randomly generated instances in a very short amount of time. However, as the correlation between the variables is increased, the difficulty of the problem increases. Recently a new class of knapsack problems was introduced by D. Pisinger [1] called the spanner knapsack instances. These instances are unsolvable by the core branch and bound instances; and as the size of the coefficients and the capacity constraint increase, the spanner instances are unsolvable even by dynamic programming based algorithms. In this paper, a genetic algorithm is presented for spanner knapsack instances. Results show that the algorithm is capable of delivering optimum solutions within a reasonable amount of computational duration.

**The 0/1 Knapsack Problem**

The 0/1 knapsack problem is defined as follows: given a set of $n$ objects each with weight $w_i$ and value (profit) $v_i$ and the capacity of the knapsack $W$

$$\text{Maximize: } \sum_{i=1}^{n} x_i . v_i \quad \text{(Equation 1)}$$

$$\text{Such that: } \sum_{i=1}^{n} x_i . w_i \leq W \quad \text{(Equation 2)}$$

$$x_i \in \{0,1\}$$

The problem is a reduction of real world resource allocation problems with constraints. The greedy algorithm described in [2] is guaranteed to produce a solution greater than half the optimum.

The 0/1 knapsack problem is weakly NP-hard in that there exist pseudo-polynomial time algorithms based on dynamic programming that can solve it exactly. There are also the core branch and bound algorithms [2] that can solve large randomly generated instances in a very short amount of time. However, as the correlation between the variables is increased, the difficulty of the problem increases. Recently a new class of knapsack problems was introduced by D. Pisinger [1] called the spanner knapsack problems. These instances are unsolvable by the core branch and bound instances; and as the size of the variables and the capacity constraint increase, the spanner instances are unsolvable even by dynamic programming based algorithms.

**The spanner 0/1 knapsack problem instances**

The spanner knapsack instances are generated from a spanner set. The spanner ($v$, $m$) instances are defined by three parameters as described by D. Pisinger [1]: $v$: the size of the spanner set; $m$: the multiplier range and the type of correlation (uncorrelated, weakly correlated and strongly correlated). We consider strongly correlated spanner (2, 10) instances in this paper, as they are the harder to solve as compared to the uncorrelated and weakly correlated counterparts.

The spanner instances are generated as described by D. Pisinger [1]: First, a set of $v$ (2 in our case) items is generated according to the distribution (strongly correlated) in the closed interval [1, $R$]. The items ($p_k, w_k$) in the spanner set are then normalized by dividing the profits and weights by $m+1$. The $n$ items are then constructed by repeatedly choosing an item ($p_k, w_k$) from the spanner set and a multiplier *a* randomly generated in the closed interval [1, *m*]. The constructed item has profit and weight ($a*p_k, a*w_k$). As described in [1], the problem instances are harder to solve for smaller spanner sets. We choose *v=2* and *m=10* (which is the same as used in [1]).

**Genetic Algorithms**

Genetic algorithms introduced in [3], gain their inspiration from the natural process of evolution, and emulate evolution by applying recombination, mutation and natural selection on a population. A meaningful building block is a low order schemata with a high potential to form a complete solution that represents the optimum. As the genetic algorithm progresses through generations, meaningful building blocks proliferate, and bad building blocks reduce from the population. Genetic algorithms have been applied to hard combinatorial optimization problems and results show that they are powerful and efficient if used judiciously.

**The Genetic Algorithm**

*Representation:* An individual of the population is represented by a string of bits such that the original order of the objects is preserved. *Mutation:* The mutation rate is adaptively changed from an initial 0.0 in steps of 1/10N till 10/N, depending on the consistent progress of the algorithm in improving upon the global best individual. If improvement is detected, the mutation rate is reset to 0.0. This technique of adapting the mutation rate, when plotted on a graph, forms an increasing magnitude saw tooth curve. *Selection:* Selection is 1-elitist. To reduce the effect that selection pressure has on premature convergence, the following mechanism is used: by default, Roulette wheel [4] is used. The appearance of a stagnant high fitness individual in the population is an indication that the search might be going in the wrong direction, and so in that case, selection is switched to random selection with a probability of 0.5. *Crossover:* The local adaptation of crossover is used as described in [5]. Crossover type for an individual is chosen randomly when the initial population is generated. When the crossover operation is performed the following algorithm is used – if the crossover type of parents is the same, use that type. Otherwise, choose at random. After crossover, the crossover types of the parents are updated according to the effectiveness of the crossover type, which is measured by the fitness difference between the offspring and the parents. A combination of greedy crossover (described later) and one point crossover is used. *Replacement:* All offspring replace their parent individuals. *Population:* A population size of 200 is used for all problems. *Generations:* The algorithm is allowed to run for 1000 generations or five minutes, whichever earlier.

The initial population is generated with the probability of choosing an object 0.7. The higher this probability, the faster the algorithm converges; however, the higher this probability, the more are the chances that the algorithm will converge around the greedy estimate. This way of generating the initial population introduces a lot of invalid solutions (violating equation 2) into the population. All invalid solutions have positive fitness, which is calculated by a linear annealing schedule [6][7]:

$$f(s) = f(gBest)/g + f'(s)/c \qquad \text{(Equation 3)}$$

$$f'(s) = W - \sum_{i=1}^{n} x_i . w_i \qquad \text{(Equation 4)}$$

Where $f'(s)$ is the negative difference between the maximum allowed weight *W* and the weight of this solution. $f(gBest)$ is the fitness of the global best, *g* is the current generation and *c* is a constant specific to a problem instance (typical value: 100), chosen small enough so that invalid solutions do not dominate the selection process. To compensate for invalid solutions, we investigated the use of a highly constructive greedy crossover. The greedy crossover takes the objects with the best $v_i/w_i$ from parents and constructs one offspring such that it is always a valid solution. (Equation 2)

As shown in table-1, the genetic algorithm is able to solve the spanner instances using a reasonable amount of computational duration for problem sizes of 100, 500, 1000, 2000 and 5000. The algorithm is run 20 times on the same instance to verify its consistency. For all but the 5000 variable instance, the algorithm is able to solve the problem everytime it is run (in 20 runs). For the 5000 variable instance, it solved the problem 17 out of 20 times, which is encouraging, as it is non-trivial for a genetic algorithm to solve a problem of this magnitude.

Table-1: Strongly Correlated Spanner (2, 10) Instances – 20 runs

| Problem | $R$ | $W/\sum w_i$ | Optimum | Greedy Estimate | Best Value | Exceed Greedy | Solve Completely | Solves completely (mean) |
|---|---|---|---|---|---|---|---|---|
| $P_{100}^{sp}$ | 5000 | 0.5 | 247079 | 246680 | - | 20\20 | 20\20 | 2.02 seconds |
| $P_{500}^{sp}$ | 5000 | 0.5 | 1255793 | 1255381 | - | 20\20 | 20\20 | 1.8 seconds |
| $P_{1000}^{sp}$ | 10000 | 0.05 | 284886 | 284204 | - | 20\20 | 20\20 | 8.4 seconds |
| $P_{2000}^{sp}$ | 10000 | 0.05 | 416021 | 415925 | - | 20\20 | 20\20 | 9.4 seconds |
| $P_{5000}^{sp}$ | 10000 | 0.05 | 1606305 | 1605825 | - | 20\20 | 17\20 | 31.03 seconds |

**Conclusion**

We presented a genetic algorithm for a class of hard knapsack problems and showed that the algorithm is capable of delivering the optimum solutions to the spanner knapsack instances in a reasonable duration of time. The algorithm is general enough to be used in other fields of optimization where a utility ratio (like the greedy estimate) is available. Results show that the algorithm solves the 0/1 knapsack spanner instances everytime it is run in 20 runs in all but the 5000 variable instance. For the 5000 variable instance, it solved the problem 17 out of 20 runs which is encouraging as it is non-trivial for a randomized algorithm to solve a problem of this magnitude.